\documentclass[letterpaper]{article} 
\usepackage{aaai25}  
\usepackage{times}  
\usepackage{helvet}  
\usepackage{courier}  
\usepackage[hyphens]{url}  
\usepackage{graphicx} 
\def\UrlFont{\rm}  
\usepackage{natbib}  
\usepackage{caption} 
\usepackage{algorithm}
\usepackage{algorithmic}
\usepackage{amssymb}
\usepackage{xcolor}
\usepackage{array}
\usepackage{colortbl}
\usepackage{graphicx}
\usepackage{multirow}
\usepackage{booktabs}
\usepackage{pifont}
\usepackage{amsmath}
\usepackage{tabularx}

\definecolor{tgreen}{RGB}{32,178,170}

\definecolor{tbg}{RGB}{230,245,230}
\definecolor{tbb}{RGB}{135,206,250}
\definecolor{blue2}{RGB}{0,139,139}
\definecolor{red2}{RGB}{255,127,80}

\newcommand{\cmark}{\color{tgreen}\ding{51}}%
\newcommand{\xmark}{\color{red2}\ding{55}}%

\usepackage{newfloat}
\usepackage{listings}
\DeclareCaptionStyle{ruled}{labelfont=normalfont,labelsep=colon,strut=off} 
\floatstyle{ruled}
\newfloat{listing}{tb}{lst}{}
\floatname{listing}{Listing}
\title{Pragmatist: Multiview Conditional Diffusion Models for \\ High-Fidelity 3D Reconstruction from Unposed Sparse Views}
\author{
    Songchun Zhang,
    Chunhui Zhao\equalcontrib
}
\affiliations{
    State Key Laboratory of Industrial Control Technology,\\
    College of Control Science and Engineering, Zhejiang University, China

}

\usepackage{bibentry}

\begin{document}

\maketitle
\begin{abstract}
Inferring 3D structures from sparse, unposed observations is challenging due to its unconstrained nature.
Recent methods propose to predict implicit representations directly from unposed inputs in a data-driven manner, achieving promising results.
However, these methods do not utilize geometric priors and cannot hallucinate the appearance of unseen regions, thus making it challenging to reconstruct fine geometric and textural details.
To tackle this challenge, our key idea is to reformulate this ill-posed problem as conditional novel view synthesis, aiming to generate complete observations from limited input views to facilitate reconstruction.
With complete observations, the poses of the input views can be easily recovered and further used to optimize the reconstructed object.
To this end, we propose a novel pipeline \textit{Pragmatist}.
First, we generate a complete observation of the object via a multiview conditional diffusion model.
Then, we use a feed-forward large reconstruction model to obtain the reconstructed mesh.
To further improve the reconstruction quality, we recover the poses of input views by inverting the obtained 3D representations and further optimize the texture using detailed input views.
Unlike previous approaches, our pipeline improves reconstruction by efficiently leveraging unposed inputs and generative priors, circumventing the direct resolution of highly ill-posed problems.
Extensive experiments show that our approach achieves promising performance in several benchmarks.

\end{abstract}
\section{Introduction}
\label{sec:intro}

\begin{figure}[t!]
    \centering
    \includegraphics[width=0.49\textwidth]{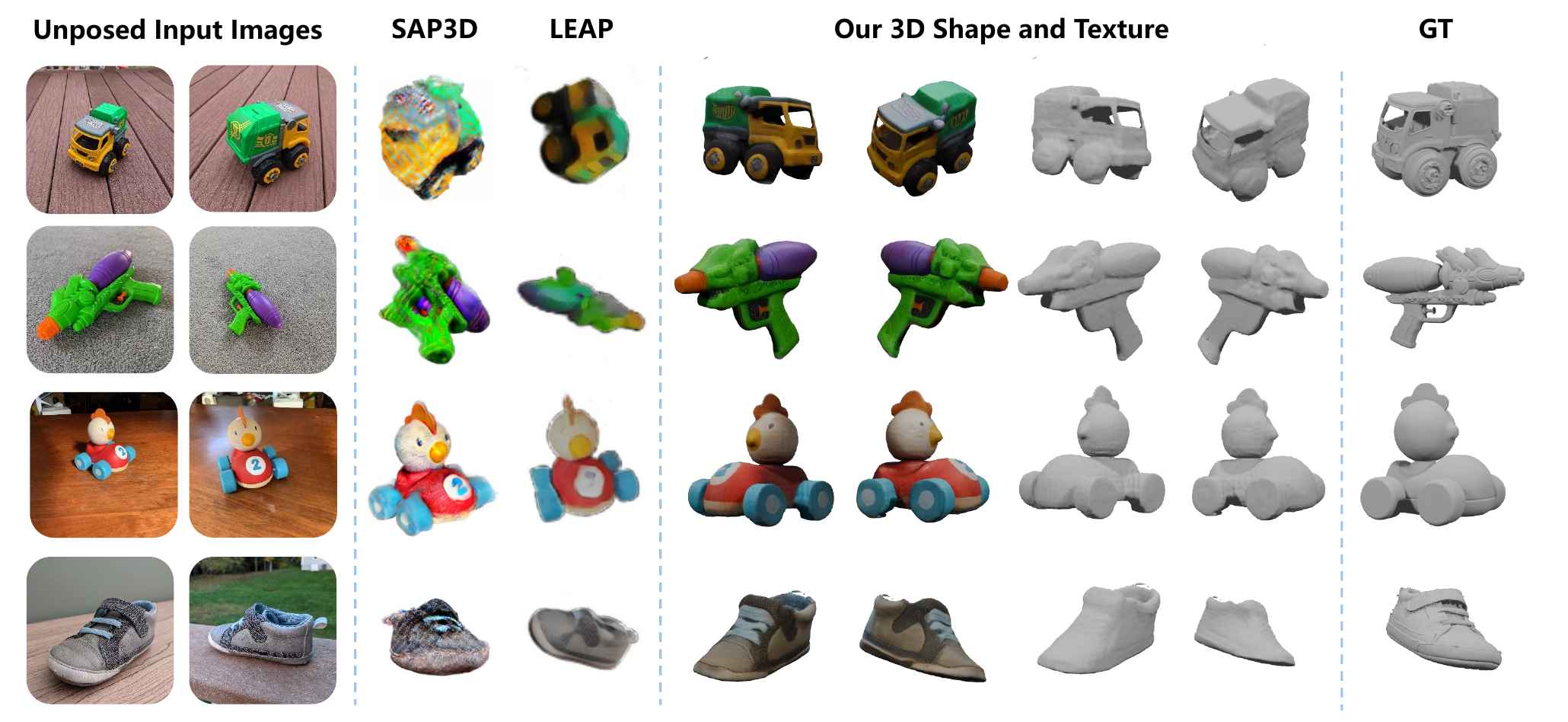}
    \caption{Our method reconstructs 3D geometry and texture from sparse unposed images, even across different scenes of the same object. More results are available in the supplementary material.}
    \label{fig:teaser}
\end{figure}

Dense multi-view 3D reconstruction~\cite{colmap, neus,barnes2009patchmatch} has always been one of the core challenges in computer vision and graphics. 
Current 3D reconstruction techniques have achieved impressive results in various applications such as film production, e-commerce, and virtual and augmented reality.
%
This paper focuses on the specific setting of reconstructing 3D objects from unposed sparse view observations, which is closer to real-world scenarios with only partial observations (e.g., casual capture and e-commerce scenarios).

This problem is particularly challenging from many perspectives, including lacking the explicit geometric information provided by the camera poses, and the need to infer the complete 3D structure from only limited partial observations.
To overcome these challenges, some methods have achieved promising results, either by pre-training models on large-scale datasets to introduce prior knowledge of 3D scenes~\cite{tang2024lgm,chen2024lara,yu2021pixelnerf,li2023instant3d,zhang2023cross}, or by optimizing scene by scene through manually designed regularization~\cite{zou2024sparse3d,zhou2023sparsefusion,wang2023sparsenerf}.
However, such methods rely heavily on the known accurate camera poses of the input image, which is usually only available in synthetic datasets or through privileged information in additional views.
To tackle this challenge, some methods~\cite{jiang2022few,jiang2023leap,kani2023upfusion,pflrm} adopt a fully feed-forward pose-free framework by learning geometric priors from large-scale 3D objects through spatial-aware attention.
However, these methods have difficulty recovering complex geometries and textures due to volumetric resolution constraints. 
In addition, due to the lack of geometric prior, it is challenging for these methods to converge on datasets containing complex objects.
Recent methods~\cite{sap3d, wu2023ifusion} first estimate the sparse camera pose from images using pre-trained models, and then reconstruct the 3D model via test-time optimization, which can achieve superior quality compared to the feed-forward methods.
However, these methods do not fully utilize the input view information, and the pose estimation error may lead to suboptimal 3D reconstruction results.

\begin{table}[t!]
\centering
\fontsize{9}{10}\selectfont
\setlength{\tabcolsep}{1mm}
\begin{tabular}{l|ccccc}
\toprule
Method &
  \begin{tabular}[c]{@{}c@{}} Unposed \\ Inputs\end{tabular} &
  \begin{tabular}[c]{@{}c@{}} 3D \\ Consist.\end{tabular} &
  Genl. &
  \begin{tabular}[c]{@{}c@{}} Gen \\ Occ. \end{tabular}&
  \begin{tabular}[c]{@{}c@{}} Lev. 2D\\ Inputs\end{tabular} \\ \midrule
SparseFusion~\shortcite{zhou2023sparsefusion} & \xmark & \cmark & \cmark & \cmark & \cmark \\
PixelNeRF~\shortcite{yu2021pixelnerf}    & \xmark & \cmark & \cmark & \xmark & \xmark \\
LEAP~\shortcite{jiang2023leap}         & \cmark & \cmark & \cmark & \xmark & \xmark \\
PF-LRM~\shortcite{pflrm}       & \cmark & \cmark & \cmark & \xmark & \xmark \\
SAP3D~\shortcite{sap3d}        & \cmark & \cmark & \xmark & \cmark & \cmark \\
UpFusion~\shortcite{kani2023upfusion}     & \cmark & \cmark & \cmark & \cmark & \xmark \\
\rowcolor[HTML]{EFEFEF}
\textbf{Ours}         & \cmark & \cmark & \cmark & \cmark & \cmark \\ \bottomrule
\end{tabular}%
\caption{\textbf{Comparison with prior methods.} \textit{Unposed Inputs} indicates whether the method can handle unposed input images. \textit{3D Consist.} refers to the method's ability to maintain 3D consistency. \textit{Genl.} denotes the generalization capability of the method. \textit{Gen Occ.} indicates whether the method can hallucinate unseen regions. \textit{Lev. 2D Inputs} denotes the method's ability to leverage 2D input images.}
\label{tab:method_compare}
\end{table}

In this paper, we propose a novel framework, named \textit{Pragmatist}, to solve the challenge of high-fidelity 3D reconstruction from unposed sparse views, as shown in Figure~\ref{fig:teaser} and Table~\ref{tab:method_compare}.
Our innovation is to transform the ill-posed 3D reconstruction problem into a conditional novel view generation task.
This is achieved by generating additional consistent observations via a multi-view conditional diffusion model, which effectively transforms the sparse view 3D reconstruction problem into a fully observed 3D reconstruction setting.
Specifically, we first propose a pose-free multi-image conditional multi-view generation model, in which the self-attention in 2D latent features can learn 3D consistency across the conditional views and generated views, without explicitly using the camera pose information of the conditional views.
In contrast to previous methods~\cite{eschernet,zero123,gao2024cat3d}, our approach does not require the camera pose of the input view or coupling the output view pose with the input view during the generation process, making it more flexible and applicable to a wider range of applications.

However, reconstructing accurate texture and geometric details directly from the generated multi-view images is quite challenging.
To address this, we propose to combine the advantages of generative priors and geometric constraints.
First, we use a feed-forward mesh generation model to reconstruct the object model from the generated target views.
Then, we estimate the position and pose of the input views within the object coordinate system via 2D-3D matching.
Finally, we further refine the texture using high-resolution input views, restoring detailed geometric structures and material textures.
%
Comprehensive evaluations show that \textit{Pragmatist} improves the reconstruction quality of sparse unposed inputs and can effectively reconstruct complex geometries and realistic textures of 3D objects. 
%
In a nutshell, our key contributions can be summed up as follows:
\begin{itemize}
    \item We propose a novel paradigm for unposed sparse views 3D reconstruction by reformulating the problem as conditional novel view synthesis, aiming to generate complete observations of the given object for reconstruction.
    \item We propose a flexible multi-view conditional generative model that produces a comprehensive set of observations in canonical coordinate, based on unposed sparse input views.
    \item To improve the fidelity of the reconstruction, we propose to first recover the pose of the input view by inverting the triplane representation. Then, the fine-grained information in the input view is used to optimize the surface texture of the object.
\end{itemize}

\begin{figure*}[t!]
    \centering
    \includegraphics[width=0.95\textwidth]{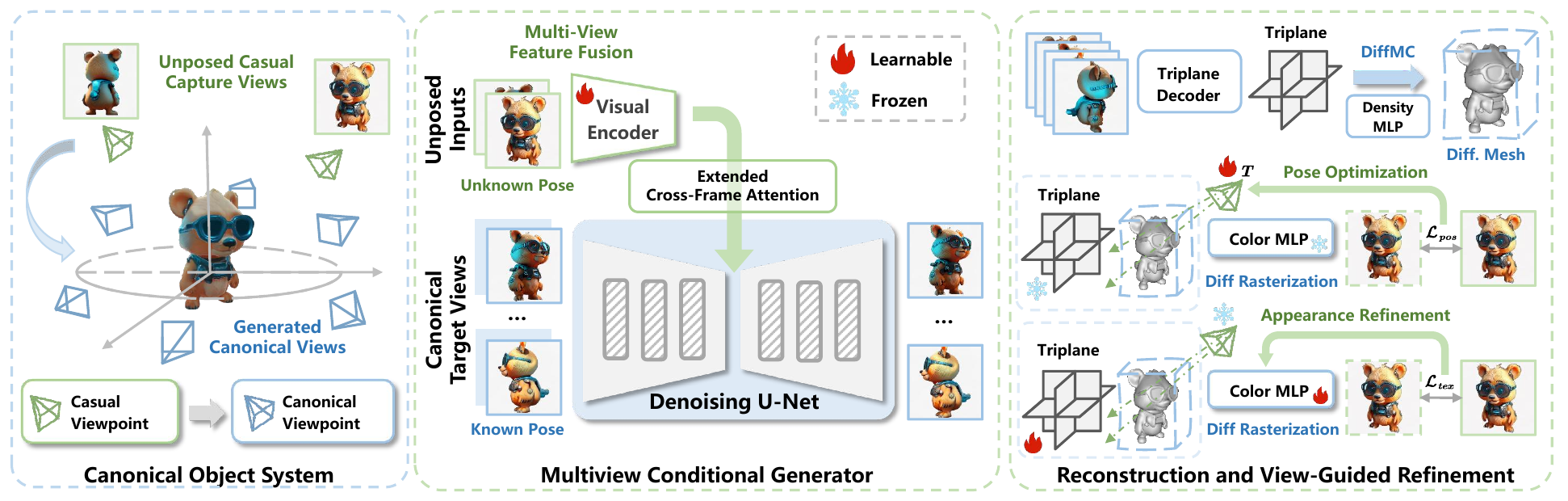}
    \caption{\textbf{Overview of our pipeline.} Our core insight is to generate additional novel views in the canonical object coordinate by conditioning with casual viewpoints to solve sparse unposed view reconstruction. The multi-view conditional generator uses a lightweight encoder to extract input view features and implicitly models the 3D consistency between the input view and the output view in the canonical coordinate system via cross-frame attention. To achieve the trade-off between the input view and the generated prior, the object mesh and triplane features are first obtained using the generated views. The input views are then aligned to the canonical coordinates via diff. rendering, and finally, the surface texture is optimized through the input views.}
    \label{fig:pipeline}
\end{figure*}
\section{Related Work}
\label{sec:related}

\noindent{\textbf{Sparse-View Pose Estimation.}}
%
Sparse-view pose estimation is a challenging task that involves determining camera poses from limited observations.
Some learning-based methods~\cite {zhang2022relpose,lin2023relposepp} use energy-based models to compose relative rotations into sets of camera poses. 
Other methods~\cite{sinha2023sparsepose, idpose} learn to refine sparse camera poses iteratively given an initial estimate, while some others~\cite{forge} exploit synthetic data to learn camera poses.
However, these methods do not generalize well to out-of-domain objects and camera viewpoints.
Additionally, some approaches~\cite{wang2023posediffusion,raydiffusion} use a diffusion model to denoise camera poses. 
Recent works predict sparse poses by predicting aligned point clouds~\cite{wang2024dust3r, pflrm} and using PnP to recover cameras.
However, these methods require large-scale synthetic data for training, and the estimated pose error can lead to suboptimal results in the subsequent reconstruction process.

\noindent{\textbf{Sparse-View Reconstruction.}}
Existing reconstruction or NVS methods perform well with dense inputs but struggle to reconstruct satisfactory results when trained on sparse observations due to insufficient information.
To address this challenge, some approaches have proposed to pretrain on large-scale datasets~\cite{yu2021pixelnerf,tang2024lgm,chen2024lara, genvs} to introduce prior knowledge.
In contrast, other approaches focus on optimizing each scenario via regularization methods~\cite{dietnerf, zou2024sparse3d, zhou2023sparsefusion, wang2023sparsenerf, yang2023freenerf}. 
Sparse3D~\cite{zou2024sparse3d} and SparseFusion~\cite{zhou2023sparsefusion} tend to distill diffusion for sparse view reconstruction.
However, these methods rely on the pose ground truth of the sparse input view, which is not practical in the real world.
To address this problem, some methods have combined sparse view pose estimation~\cite{sinha2023sparsepose, raydiffusion, lin2023relposepp} with advanced reconstruction methods, or considered methods that jointly optimize poses together with reconstruction~\cite{forge, sap3d, wu2023ifusion}.
However, the computation of explicit poses is not always reliable, leading to poor reconstruction performance.
Some methods~\cite{jiang2023leap, kani2023upfusion} use a fully pose-free model structure, but have difficulty reconstructing complex geometries and textures.

\noindent{\textbf{Multi-View Diffusion Models.}}
%
Previous 3D generative reconstruction methods based on 2D diffusion models have poor 3D generative quality due to the lack of 3D prior knowledge.
Zero123~\cite{zero123} infused 3D prior knowledge into the 2D diffusion model by pre-training on large-scale synthetic 3D data, but multi-view output still cannot guarantee consistency.
To solve this problem, many subsequent works~\cite{liu2023syncdreamer,long2024wonder3d,shi2023MVDream,voleti2024sv3d,chen2024v3d,kant2024spad,tang2024mvdiffusion++,tang2023MVDiffusion,free3d} focused on improving the 3D consistency of multi-view images.
These methods enhance the quality of generated multi-view images by modeling 3D generation through a joint probability distribution.
However, when single-view inputs are of low quality, they can result in implausible shapes and textures.
Therefore, some methods~\cite{eschernet,tang2024mvdiffusion++,gao2024cat3d,wu2024reconfusion} incorporate flexible multi-view image conditioning into the diffusion model to achieve more controllable 3D object reconstruction.
However, most of them require the ground truth camera pose (which is challenging to acquire in practice) and have specific requirements for the range of viewpoints and image quality of the input images.


\section{Method}
\label{sec:method}

Given a set of sparse observations of an object, our task is to recover the shape and texture of the current object.
%
Our method incorporates generative priors with geometric knowledge to alleviate the ambiguity of the ill-posed problem by generating additional novel views in canonical coordinates.
The overview of our pipeline is illustrated in Figure~\ref{fig:pipeline}.
In this section, we first describe the construction of a conditional multiview generator that converts arbitrary unposed sparse views into multiview images with 3D consistency.
Next, we will introduce the design of the feed-forward reconstruction model, which uses the generated images to reconstruct high-quality meshes.
Finally, we will explain how to efficiently recover the camera pose of the input observation through the reconstruction results and further improve the quality of the 3D model.

\subsection{Multiview Conditional Multiview Generator}
Existing multi-view diffusion methods usually assume that the intrinsics of the input and generated images are the same.
However, in practice, the input images can come from arbitrary cameras, and when the intrinsics are quite different from the training dataset, the generated multi-view images and reconstructed 3D meshes will be distorted.
In addition, these methods usually support only a single image in the near-orthogonal view or multi-view with ground truth camera poses as the condition, which greatly limits the application of the model in real-world scenarios.
In contrast, our model can receive arbitrary numbers of unposed inputs and generate novel views in the canonical coordinate system for subsequent reconstruction steps.
Specifically, given a sparse set of $ N $ unposed images ${\mathcal{I}}_{cond} = \{ \boldsymbol{I}_i^{cond} \}_{i=1}^N $ and their corresponding camera poses $\mathcal{P}_{cond}$, our multiview diffusion model can generate multiple novel views $\mathcal{I}_{tgt} = \{ \boldsymbol{I}_j^{tgt} \}_{j=1}^M $ based on given target camera poses $\mathcal{P}_{tgt} = \{ \boldsymbol{p}_j \}_{j=1}^M $.  
This process can be formulated as:
\begin{equation}
    \mathcal{I}_{tgt} \sim p(\mathcal{I}_{tgt} | \mathcal{I}_{cond}, \mathcal{P}_{cond}, \mathcal{P}_{tgt}),
\end{equation}
where $\mathcal{I}_{cond}$ and $\mathcal{P}_{cond}$ are the observed images and their corresponding poses, while $\mathcal{P}_{tgt}$ specifies the camera poses for the novel views to be generated.

\noindent{\textbf{Conditioning Feature Extraction.}}
Previous methods use the frozen CLIP ViT and denoising Unet to capture the high-level and low-level information of inputs, respectively, but are only suitable for single-view input.
Instead, we consider directly integrating the high- and low-level signals of multiple input views using a conditional encoder and representing them as a set of tokens for subsequent injection into a diffusion model.
Specifically, we chose to fine-tune a lightweight encoder $\mathcal{E}_{cond}$~\cite{woo2023convnext} to integrate image features and use these features to construct conditional tokens.
\\
\noindent{\textbf{Multiview Conditioned Generation.}}
We extend the generative network based on the origin latent diffusion architecture.
Specifically, the self-attention layer in the original stable diffusion is used to learn relationships within a single image, which we extend to learn relationships between multiple target views to ensure inter-view consistency.
In addition, we also extend the cross-attention layer to learn the interactions between input and target views, thereby incorporating the information of unposed input views into the target view generation process.
To enhance camera control ability, previous methods~\cite{li2023instant3d,hong2023lrm} encode camera poses at the frame level, sharing the same position encoding between pixels.
Aiming at more accurate pixel-level encoding, we integrate the camera conditions into the denoising network by parameterizing the rays $\boldsymbol{r} = (\boldsymbol{d}, \boldsymbol{o} \times \boldsymbol{d})$ and employing a two-layer MLP that injects a Plücker ray embedding for each pixel.

\noindent{\textbf{Canonical Camera System.}}
%
Generating novel view images in canonical camera settings is challenging due to the sparse input view and arbitrary viewpoint, which require the model to infer the elevation and the object's scale. 
To address these issues, we adopt a fixed absolute elevation angle and distance, using the relative azimuth angle as the pose for the new view. 
This approach eliminates uncertainties in direction and scale, removing the need for additional elevation angle estimation. 
Cameras are positioned at a distance of 1.5 with azimuth angles $(\beta, \beta \pm 45^\circ, \beta \pm 90^\circ, \beta \pm 135^\circ, \beta \pm 180^\circ)$ at 0° elevation.

\noindent{\textbf{Training Details.}}
During training, the number of condition and target images are fixed, but the inference can support any number of images.
In addition, to simulate the real-world conditional images, we perform noise perturbation and resize operations on the condition images during training to make our model more robust to the input views.

\subsection{Feed-forward Reconstruction}
In our model, the feed-forward reconstruction module aims to efficiently reconstruct high-quality meshes from images generated by the multi-view conditional generator.
However, direct optimization based on the mesh representation is challenging, as it is sensitive to geometry initialization and can get stuck in local minima, especially when reconstructing high-resolution meshes.
To solve this problem, we first optimize the NeRF representation by volume rendering, which is used to initialize the network weights for mesh reconstruction. Then, we further optimize the surface rendering loss by differentiable rasterization, which further improves the mesh surface extraction results.
\\
\noindent{\textbf{Volume Rendering Training.}}
We use a transformer-based model to predict the triplane NeRF in a feed-forward manner from sparse view inputs. 
The model consists of an image encoder, an image-to-triplane decoder, and a NeRF decoder.
The image encoder converts the multi-view images $\{I_i\}^M_{i=1}$ into latent codes.
Then, we use concatenated image codes in the image-to-triplane decoder to predict the triplane features $\boldsymbol{T}$ of the 3D object.
Unlike previous methods~\cite{hong2023lrm,li2023instant3d}, we use two separate MLPs to estimate color and density, because the density MLP $f_d$ and color MLP $f_c$ are used separately during mesh extraction and surface rendering, and can increase the stability of the training process.
The loss for volume rendering training is formulated as:
\begin{equation}
    \mathcal{L}_{vol} = \mathcal{L}_{rgb} + \lambda_p \mathcal{L}_{LPIPS},
\end{equation}
where $\mathcal{L}_{rgb}$ indicates L2 regression loss of RGB image, $\mathcal{L}_{LPIPS}$ denotes the perceptual loss~\cite{lpips}, and $\lambda$ indicates the loss weight.
\\
\noindent{\textbf{Surface Rendering Training.}}
Although the trained model shows promising results in novel view synthesis, directly extracting meshes from the density field produces artifacts.
Therefore, we propose to fine-tune the network using differentiable marching cubes (DiffMC)~\cite{diifmc} and differentiable rasterization~\cite{Laine2020diffrast}.
We first compute the density grid using triplane features and extract grid surfaces from it using the DiffMC technique.
Then, we use the differentiable rasterizer to render novel views from the extracted grid using triplane features, calculating the rendering loss to optimize the model for end-to-end mesh reconstruction.
Specifically, we first obtain the 3D position of each pixel via the differentiable rasterizer, then query the corresponding triplane feature, and use $f_c$ to obtain the color of each pixel.
The pipeline is fine-tuned using MSE and LPIPS~\cite{lpips} losses on images, MSE losses on depths and masks, and a regularization loss on opacity~\cite{wei2024meshlrm}.
The loss for surface rendering training is formulated as:
\begin{equation}
    \mathcal{L}_{geo} = \mathcal{L}_{rgb} + \lambda_{p} \mathcal{L}_{LPIPS} + \lambda_{d} \mathcal{L}_{d} + \lambda_{m} \mathcal{L}_{m} + \lambda_{r} \mathcal{L}_{o},
\end{equation}
where $\lambda_p$, $\lambda_d$, $\lambda_m$ and $\lambda_r$ denote the weights of each loss term, respectively.
Since mesh rasterization is very efficient, we render the image at full resolution for supervision.

\begin{figure*}[t!]
    \centering
    \includegraphics[width=0.95\textwidth]{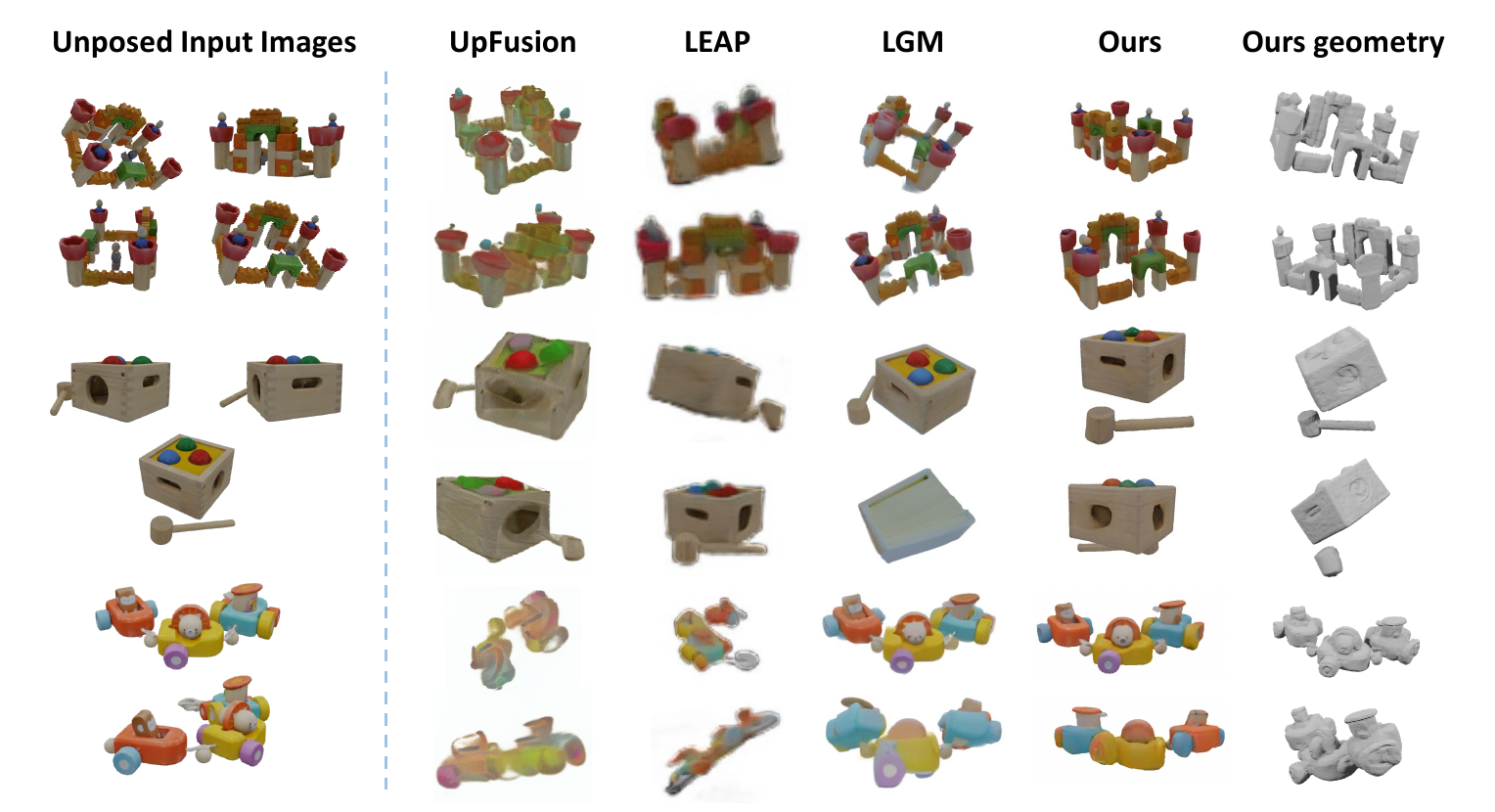}
    \caption{Qualitative comparison of unposed sparse-view 3D reconstruction results on the GSO dataset. Our method can produce more detailed and higher-quality reconstruction results than the benchmark method and can support different numbers of unposed inputs. More results can be found in the supplementary material.}
    \label{fig:multiview}
\end{figure*}

\subsection{Directing Virtual Camera for Refinement}
Though the 3D reconstruction quality of the feed-forward pipeline is promising, it still has trouble handling complex textures.
This issue gets even worse when using unposed images as input.
We argue that the rich high-resolution texture information in the input unposed view should be fully utilized.
Inspired by the previous incremental SFM pipeline~\cite{colmap}
we can treat these unposed views as new inputs and register them into the reconstruction through visual relocalization, thereby optimizing the reconstruction.

\noindent{\textbf{Align Camera to Canonical Coordinate.}}
To achieve this, using the mesh from feed-forward reconstruction and the camera pose, we can obtain the corresponding image by mesh rendering.
Subsequently, we use the photometric loss for backpropagation to optimize the camera pose $\boldsymbol{p}$ of $\boldsymbol{I}^{cond}$.
In practice, objects are usually not centered on the screen. 
To align the input view with the canonical coordinate system, we use foreground segmentation to obtain the object mask and then generate an object-centered image $\boldsymbol{I}^{cond}_v$ via homography transformation.
We then use the spherical coordinate system $\boldsymbol{p}=\{\alpha,\beta,\gamma,r\}$ to describe the input camera poses.
This simplified representation reduces the search space in pose optimization.
The process can be formulated as the following equation:
\begin{equation}
    \hat{\boldsymbol{p}} = \operatorname*{arg\,min}_{\boldsymbol{p}} 
    ~\mathcal{L}_{pos}(\mathcal{R}(\boldsymbol{T}, \boldsymbol{p}, f_c, f_d), {I}^{cond}_v)
\end{equation}

During pose optimization, the parameters of the $\boldsymbol{T}$, $f_c$, and $f_d$ are frozen.
To avoid local minima due to poor initial values, we start with different initial poses and choose the one with the minimum loss after convergence as the result.
\\
\noindent{\textbf{View-Guided Appearance Refinement.}}
To improve the fidelity of 3D reconstruction models to input views, we propose to use input images that have been registered to canonical coordinates to further optimize the texture of the model.
Specifically, we use the predicted poses of the input views to obtain the features of the 3D surface points via the differentiable rasterizer and the obtained triplane feature $\boldsymbol{T}$, and then use $f_c$ to render current views.
Then, we calculate the loss between the render views and the input views to optimize the color MLP and the triplane feature $\boldsymbol{T}$, thereby improving the appearance of the model.
The process can be formulated as follows
\begin{equation}
    \min_{f_c,\boldsymbol{T}}~\mathcal{L}_{tex}(\mathcal{R}(\boldsymbol{T}, \hat{\boldsymbol{p}}, f_c, f_d), \boldsymbol{I}^{cond}_v).
\end{equation}
We found that separating the density and color models helped to refine the appearance without affecting the performance of the geometry model, as we only needed to fine-tune the color MLP $f_c$.

\section{Experiments}
\label{sec:exp}

\subsection{Experimental Settings}
\noindent{\textbf{Implementation details.}}
Our model is optimized using the AdamW~\cite{adamw} with a weight decay of 0.01.
The multiview generator is trained with an initial learning rate of $3e-4$, which decreases to $1e-5$ over 30,000 training steps using a cosine learning rate decay schedule. 
There are two stages to the training of the feedforward reconstruction model. The first stage trains at \(256 \times 256\) resolution using the AdamW optimizer (\(\beta_2 = 0.95\)) with a weight decay of \(0.05\), a cosine learning rate decay of \(4 \times 10^{-4}\), and \(2500\) warm-up iterations, for a total of \(60k\) training iterations. We use a batch size of \(192\) and sample \(128\) points per ray.
The refinement stage operates at \(512 \times 512\) resolution with \(256\) ray samples and a batch size of \(64\). The learning rate is set to \(1 \times 10^{-5}\) with \(1000\) warm-up iterations, using DiffMC for mesh optimization over \(10k\) iterations.
Loss weights are \(\lambda_p = 0.5\), \(\lambda_d = 0.5\), \(\lambda_m = 1\), and \(\lambda_n = 1\). For stability, normal loss is added after convergence with the generator frozen. 
We randomly sample \(4\) input views with \(4\) supervision views during the NeRF training phase, expanding to \(8\) supervision views in the geometry refinement phase. The backbone uses a pre-LN transformer with \(24\) layers (\(1024\) channels, \(16\) attention heads, \(64\) dimensions per head). GeLU-activated MLPs with \(4096\) intermediate features split into density (3 layers) and color (4 layers) pathways, each using \(512\) hidden dimensions.


\begin{figure}[t!]
    \centering
    \includegraphics[width=0.46\textwidth]{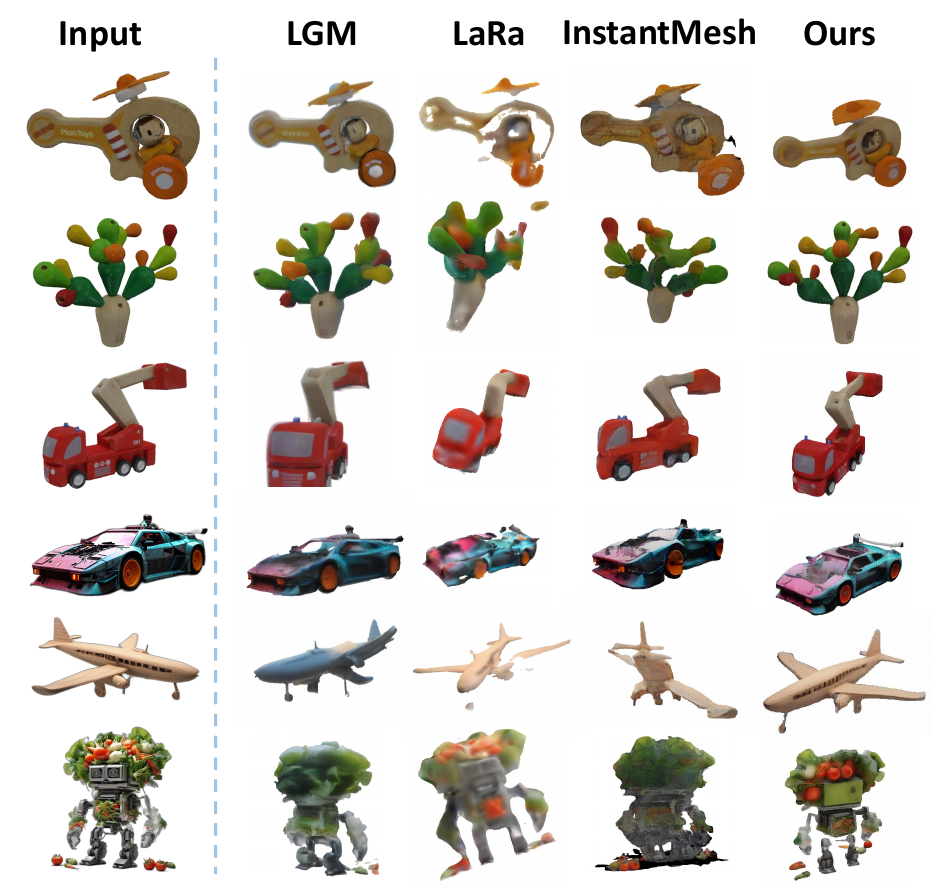}
    \caption{Qualitative comparison of single view 3D reconstruction results on the GSO and in-the-wild datasets. Though our method is not designed for single-view setting, it produces results comparable to these benchmarks and generates reasonable textures in the invisible areas.}
    \label{fig:singleview}
\end{figure}




\noindent{\textbf{Datasets.}}
Our model is trained on a filtered subset of Objaverse with about 240k quality assets.
Following the previous methods, we evaluate our model's performance on the Google Scanned Object (GSO)~\cite{gso} and OmniObject3D~\cite{wu2023omniobject3d} datasets.
To further verify the generalizability of the method, we also evaluated its performance on casually captured images and images created using a text-to-image diffusion model.


\noindent{\textbf{Metrics.}}
We evaluate our method on novel view synthesis and reconstruction tasks.
Specifically, we use PSNR, SSIM, and LPIPS as novel view synthesis metrics, while Chamfer Distance (CD) and Volume IOU are used for 3D reconstruction evaluation.
For 3D geometry evaluation, we align the generated meshes with the ground-truth meshes at the origin, rescale them to a unit cube, and apply ICP registration for alignment.


\begin{table}[t!]
\centering
\fontsize{9}{10}\selectfont
\setlength{\tabcolsep}{1mm}
\begin{tabular}{l|c|ccc|ccc}
\toprule
\multirow{2}{*}{Method}    & \multirow{2}{*}{Pose} & \multicolumn{3}{c|}{GSO} & \multicolumn{3}{c}{OmniObject3D} \\ \cmidrule(l){3-8} 
                       &      & \cellcolor[HTML]{F0FBEF}PSNR  & \cellcolor[HTML]{F0FBEF}SSIM & \cellcolor[HTML]{FEF1F1}LPIPS & \cellcolor[HTML]{F0FBEF}PSNR  & \cellcolor[HTML]{F0FBEF}SSIM  & \cellcolor[HTML]{FEF1F1}LPIPS \\ \midrule
\multirow{2}{*}{PixelNeRF} & Pred.                  & 11.483  & 0.852  & 0.216 & 10.183     & 0.847    & 0.226    \\
                       & GT   & 14.278 & 0.932 & 0.153  & 11.365 & 0.852 & 0.219  \\
\multirow{2}{*}{FORGE} & Pred. & 11.432 & 0.754 & 0.762   & 10.164 & 0.849 & 0.228  \\
                       & GT   & 15.164 & 0.867 & 0.191  & 12.456 & 0.856 & 0.211  \\
\multirow{2}{*}{LGM}   & Pred. & 20.149 & 0.888 & 0.156  & 17.698 & 0.878 & 0.173  \\
                       & GT   & 21.783 & 0.873 & 0.214  & 18.546 & 0.883 & 0.172  \\
\multirow{2}{*}{LaRa}  & Pred. & 20.696 & 0.889 & 0.152  & 17.877 & 0.878 & 0.175  \\
                       & GT   & 23.374 & 0.865 & 0.208  & 20.186 & 0.888 & 0.157  \\ \midrule
SAP3D                  & Pred. & 17.312 & 0.863 & 0.173  & 15.534 & 0.871  & 0.187  \\
LEAP                   & \xmark    & 21.348 & 0.856 & 0.128  & 19.439 & 0.885 & 0.162  \\
UpFus.               & \xmark    & 20.924 & 0.901 & 0.147  & 18.281 & 0.881 & 0.169  \\ \midrule
Ours                   & \xmark    & 22.931 & 0.887 & 0.121   & 20.897 & 0.892  & 0.153  \\
\rowcolor[HTML]{EFEFEF}
Ours & Pred. & \textbf{23.698} & \textbf{0.902} & \textbf{0.131} & \textbf{21.764} & \textbf{0.895} & \textbf{0.146}  \\ \bottomrule
\end{tabular}%
\caption{Comparison with sparse-view reconstruction methods on novel view synthesis. We conducted experiments on two datasets, i.e., GSO~\cite{gso} and OmniObject3D~\cite{wu2023omniobject3d}. We denote reconstructions using predicted poses as \textit{Pred.}, those using ground truth poses as \textit{GT}, and reconstructions without poses as \xmark.}
\label{tab:sparse_view}
\end{table}

\begin{table}[t!]  
\centering
\fontsize{9}{10}\selectfont
\setlength{\tabcolsep}{1mm}
\begin{tabular}{l|cccc}  
\toprule
{Method} & \cellcolor[HTML]{FEF1F1}{R. error$\downarrow$} & \cellcolor[HTML]{F0FBEF}{Acc.@15$^\circ$$\uparrow$} & \cellcolor[HTML]{F0FBEF}{Acc.@30$^\circ$$\uparrow$} & \cellcolor[HTML]{FEF1F1}{T. error$\downarrow$} \\
\midrule  
FORGE & 98.471 & 0.024 & 0.036 & 1.122 \\
HLoc & 95.243 & 0.028 & 0.146 & 1.189 \\
RelPose++ & 67.580 & 0.269 & 0.415 & 0.893 \\
\rowcolor[HTML]{EFEFEF}
\textbf{Ours} & \textbf{2.182} & \textbf{0.961} & \textbf{0.978} & \textbf{0.039} \\
\bottomrule
\end{tabular}
\caption{Comparison of the accuracy of pose estimation on the GSO dataset.}
\label{tab:pose}
\end{table}

\subsection{Unposed Sparse View Reconstruction}
\noindent{\textbf{Baselines.}}
We first compare our method with fully pose-free reconstruction methods, i.e., LEAP~\cite{jiang2023leap}, UpFusion~\cite{kani2023upfusion}, and FORGE~\cite{forge}, as they can directly receive sparse unposed inputs.
In addition, many recent sparse reconstruction methods have achieved promising results, but require ground truth pose as input, such as LGM~\cite{tang2024lgm}, SAP3D~\cite{sap3d}, SparseNeuS~\cite{long2022sparseneus}, and LaRa~\cite{chen2024lara}.
To compare with them, we first estimate the camera poses using the pre-trained sparse pose estimators~\cite{lin2023relposepp, raydiffusion}, and then use these methods to reconstruct to compare with our results.
\\
\noindent{\textbf{Comparison Results.}}
Compared to pose-free methods, the metrics of our method are significantly improved, as shown in Figure~\ref{fig:multiview} and Table~\ref{tab:sparse_view}, due to 1) avoiding a direct data-driven solution to the ill-posed problem, and 2) reconstructing in a canonical coordinate system, which alleviates the problem of scale ambiguity.
In addition, our method performs better than the method that predicts the camera pose in image space and then performs multiview reconstruction. The pose estimation accuracy of such methods is prone to degradation when encountering uncommon input viewpoints, which leads to cumulative errors and affects the reconstruction quality.
More results can be found in the supplementary material.
\\
\noindent{\textbf{Pose Estimation Results.}}
As shown in Table~\ref{tab:pose}, the accuracy of our method is significantly better than that of the method that directly predicts the camera pose from the image space, and some traditional relative pose estimation methods.
Our performance stems from predicting the camera pose from the obtained 3D shape rather than directly from the 2D image. 
Due to poor visual correspondence, traditional SfM methods tend to fail when dealing with sparse views, so we use the more robust HLoc as a benchmark for comparison.

\subsection{Single View Reconstruction}
\noindent{\textbf{Baselines.}}
Existing single-image 3D reconstruction methods have achieved promising results, and can be divided into two main branches, namely those that rely on generative priors~\cite{zero123, long2024wonder3d, liu2023syncdreamer, wang2024crm, tang2024lgm} and those that do not~\cite{hong2023lrm, tochilkin2024triposr}.
We compare our method with these two types of methods. To ensure fairness, we use only a single image as input for our method.
\\
\noindent{\textbf{Comparison Results.}}
As shown in Table~\ref{tab:single-view} and Figure~\ref{fig:singleview}, our method can achieve comparable performance to the current SOTA single image-to-3D method.
These methods suffer from significant performance degradation in in-the-wild scenes, mainly due to the uncertainty of the input camera view.
Previous methods often rely on existing generative models that require a range of input viewpoints, resulting in lower-quality multi-view generation and suboptimal reconstruction when encountering unusual viewpoints.
In contrast, our method is robust to the input image view and can use any multi-view inputs, increasing controllability and improving generation and reconstruction results.
More results can be found in the supplementary material.

\begin{figure}[t!]
    \centering
    \includegraphics[width=0.48\textwidth]{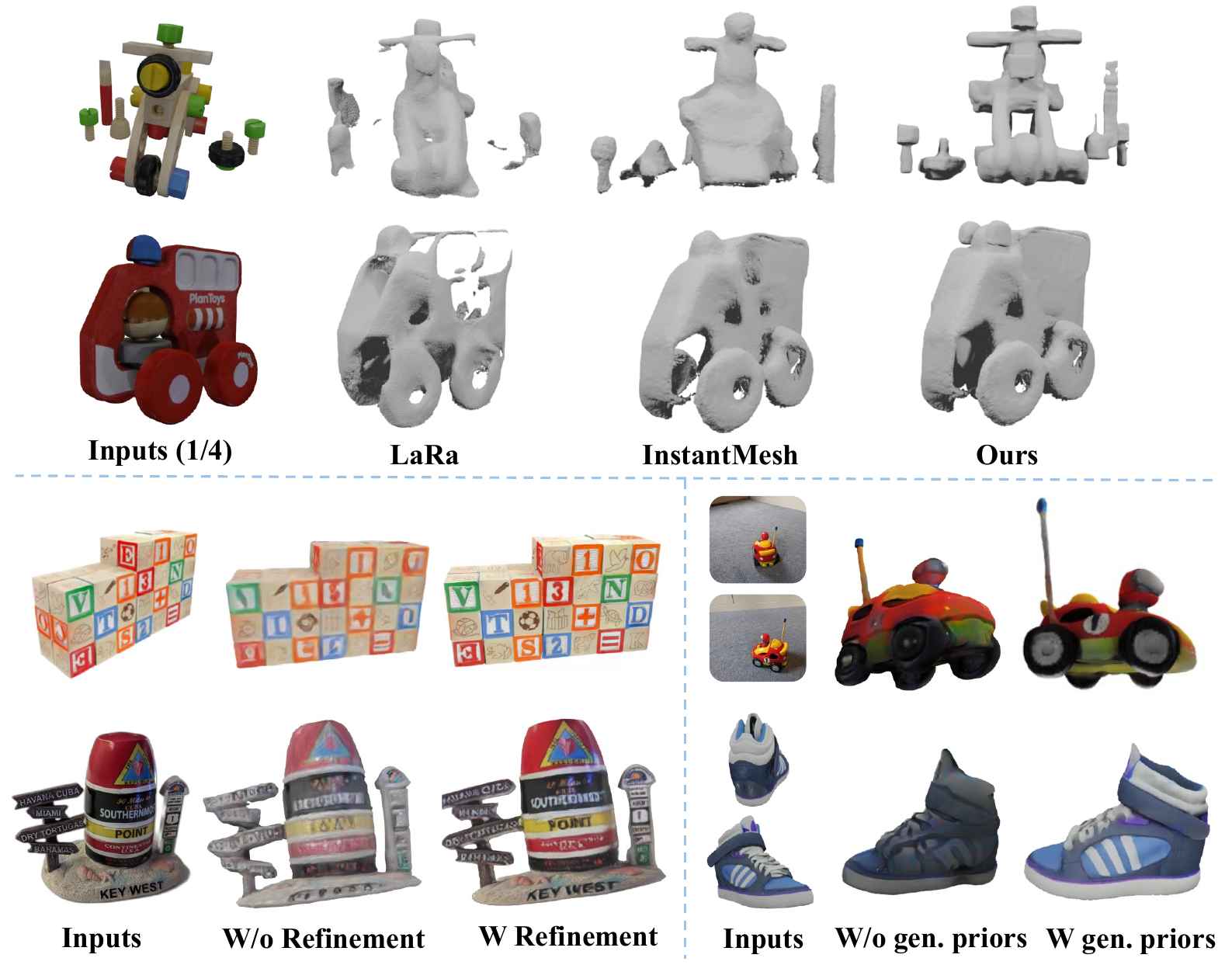}
    \caption{\textbf{Ablation Studies.} The top image illustrates the effectiveness of our reconstruction model, the bottom left image shows the effect of our refinement step, and the bottom right image demonstrates the effect of generative priors.}
    \label{fig:ablation_generative}
\end{figure}
\begin{table}[t!]
\centering
\begin{tabular}{l|c|cc|cc}
\toprule
\multirow{2}{*}{Method} &
  \multirow{2}{*}{Pose} &
  \multicolumn{2}{c|}{GSO} &
  \multicolumn{2}{c}{OmniObject3D} \\ \cmidrule(l){3-6} 
                      &      &  \cellcolor[HTML]{FEF1F1}CD$\downarrow$   & \cellcolor[HTML]{F0FBEF}IoU$\uparrow$    & \cellcolor[HTML]{FEF1F1}CD$\downarrow$     & \cellcolor[HTML]{F0FBEF}IoU$\uparrow$    \\ \midrule
\multirow{2}{*}{LGM}  & Pred. & 0.0459 & 0.2903 & 0.0532 & 0.4423 \\
                      & GT   & 0.0251 & 0.5632 & 0.0305 & 0.5843 \\
\multirow{2}{*}{LaRa} & Pred. & 0.0448 & 0.3764 & 0.0511 & 0.4312 \\
                      & GT   & 0.0246 & 0.5624 & 0.0261 & 0.5425 \\
SAP3D                 & Pred. & 0.0364 & 0.5346 & 0.0423 & 0.2876 \\ \midrule
\multirow{2}{*}{Ours} &
  \xmark &
  \multirow{2}{*}{\textbf{0.0187}} &
  \multirow{2}{*}{\textbf{0.6964}} &
  \multirow{2}{*}{\textbf{0.0202}} &
  \multirow{2}{*}{\textbf{0.6693}} \\
                      & Pred. &        &        &        &        \\ \bottomrule
\end{tabular}%
\caption{Comparison with sparse-view reconstruction methods on the reconstructed mesh. We conducted experiments on two datasets, i.e., GSO~\cite{gso} and OmniObject3D~\cite{wu2023omniobject3d}.}
\label{tab:my-table}
\end{table}
\begin{table}[t!]
\centering
\fontsize{9}{11}\selectfont
\setlength{\tabcolsep}{1.2mm}
\begin{tabular}{l|cc|ccc}
\toprule
Method         & \cellcolor[HTML]{FEF1F1}CD$\downarrow$ & \cellcolor[HTML]{F0FBEF}IoU$\uparrow$ & \cellcolor[HTML]{F0FBEF}PSNR$\uparrow$  & \cellcolor[HTML]{F0FBEF}SSIM$\uparrow$ & \cellcolor[HTML]{FEF1F1}LPIPS$\downarrow$ \\ \midrule
One-2-3-45*    & 0.0629         & 0.4086    & -      & -     & -      \\
SyncDreamer*   & 0.0261         & 0.5421    & 19.512 & 0.847 & 0.188  \\
Wonder3D*      & 0.0329         & 0.5768    & -      & -     & -      \\
LGM*           & 0.0242         & 0.5618    & 21.418 & 0.861 & 0.214  \\
CRM*           & 0.0212         & 0.6146    & 21.984 & 0.876 & 0.193  \\
TripoSR        & 0.0209         & 0.6369    & 22.376 & 0.885 & 0.131  \\
InstantMesh* &
  0.0187 &
  0.6417 &
  22.879 &
  0.901 &
  0.116 \\
Era3D*         & 0.0175         & 0.6654    & -      & -     & -      \\ \midrule
Ours* [1 View] & 0.0189         & 0.6404    & 20.455 & 0.857 & 0.156  \\
\rowcolor[HTML]{EFEFEF}
Ours* [2 Views] &
  0.0165 &
  0.6949 &
  22.763 &
  0.898 &
  0.123 \\
\rowcolor[HTML]{EFEFEF}
Ours* [4 Views] &
  \textbf{0.0146} &
  \textbf{0.7348} &
  \textbf{24.074} &
  \textbf{0.904} &
  \textbf{0.112} \\ \bottomrule
\end{tabular}%
\caption{Comparison with single-view reconstruction methods on the reconstructed mesh. We conducted experiments on the GSO dataset. '*' denotes models that employ generative priors for 3D reconstruction.}
\label{tab:single-view}
\end{table}

\subsection{Ablation Studies}
\noindent{\textbf{Ablation of Multiview Generator.}}
We first evaluate the multi-view generator in multi-view conditions. 
As shown in Figure~\ref{fig:ablation_generative}, when this module is removed, our method degenerates to pose-free reconstruction.
Even when combined with pose estimation for end-to-end reconstruction, the results are still unsatisfactory, and the texture in the unseen region is blurred due to the lack of generative priors.
More results can be found in the supplementary material.

\noindent{\textbf{Ablation of Feed-Forward Reconstruction.}}
We evaluated the effect of our multi-view reconstruction model. 
We compared our method with several other existing end-to-end reconstruction methods. 
As shown in Figure~\ref{fig:ablation_generative}, our end-to-end reconstruction method can achieve reconstruction quality comparable to the optimization method with high efficiency. In addition, we can use differentiable marching cube to quickly extract higher-quality meshes.

\noindent{\textbf{Ablation of Virtual Camera Guided Refinement.}}
We evaluated the effectiveness of the refinement step.
As shown in Figure~\ref{fig:ablation_generative}, after removing the refinement step, the reconstruction of complex textured objects becomes worse.
During the refinement process, the use of an independent MLP can improve the reconstruction performance while maintaining geometric quality.
In addition, the experimental results show that optimizing the triplane feature and color MLP at the same time achieves better results than optimizing the color MLP alone during the appearance refinement process.

\section{Conclusions}
\label{sec:conclusion}

In this paper, we introduced \textit{Pragmatist}, a novel pipeline for inferring 3D structures from sparse, unposed observations by reformulating the problem as a task of conditional novel view synthesis.
Our approach uses a multi-view conditional diffusion model to generate complete object observations, which are then used by a large reconstruction model to generate high-quality meshes.
Furthermore, we enhance reconstruction accuracy by recovering input view poses and optimizing texture details using available input views.
Extensive experiments demonstrate that \textit{Pragmatist} significantly outperforms existing methods in several benchmarks.
By utilizing unposed inputs and generative priors, our pipeline can efficiently solve this ill-posed problem and ensure generalizability across diverse scenes.


\section*{Acknowledgments}
This research was supported by the National Natural Science Foundation of China (No. 62450020) and the National Natural Science Foundation of China (No. 62125306).

\bibliography{reference}


\appendix
\section{A. Supplementary Material}

\subsection{A.1 Technical Details}
\subsubsection{Neural Radiance Fields (NeRF)}
NeRF is a novel view synthesis technique that has shown impressive results.
It represents the specific 3D scene via an implicit function, denoted as $f_\theta:(\boldsymbol{x}, \boldsymbol{d}) \mapsto (\mathbf{c}, \sigma)$, given a spatial location $\mathbf{x}$ and a ray direction $\mathbf{d}$, where $\theta$ represents the learnable parameters, and $\mathbf{c}$ and $\sigma$ are the color and density.
To render a novel image, NeRF marches a camera ray $\mathbf{r}(t) = \mathbf{o} + t\mathbf{d}$ starting from the origin $\mathbf{o}$ through each pixel and calculates its color $\hat{\boldsymbol{C}}$ via the volume rendering quadrature:
\[
 \hat{\boldsymbol{C}}(\mathbf{r})=\sum_{i=1}^N T_i\left(1-\exp \left(-\sigma_i \delta_i\right)\right) \mathbf{c}_i
\]
where $T_i=\exp \left(-\sum_{j=1}^{i-1} \sigma_j \delta_j\right)$, $\alpha_i = \left(1-\exp \left(-\sigma_i \delta_i\right)\right)$, and $\delta_{k}=t_{k+1}-t_k$ indicates the distance between two point samples. 
Typically, stratified sampling is used to select the point samples $\{t_i\}_{i=1}^N$ between $t_n$ and $t_f$, which denote the near and far planes of the camera.
When multi-view images are available, $\theta$ can be easily optimized with the MSE loss:
\[
\mathcal{L}=\sum_{\boldsymbol{r} \in \mathcal{R}}\left\|\boldsymbol{\hat{C}}(\boldsymbol{r})-\boldsymbol{C}(\boldsymbol{r})\right\|_2^2
\]

\noindent 
where $\mathcal{R}$ is the collection of rays, and $\boldsymbol{C}$ indicates the ground truth color.

\subsubsection{Large Reconstruction Model (LRM)}
LRM is a method for predicting 3D models from a single image.
It uses a pre-trained vision transformer DINO as an image encoder to extract image features, and then uses cross-attention to project 2D image features onto a 3D triplane representation, and uses self-attention to model the relationship between triplane tokens.
The output tokens are reshaped and upsampled to the final triplane feature map $\mathbf{T}$, and then volume rendering is performed through a MLP to obtain color and density $\sigma$.
After obtaining the rendered views, the model is supervised by MSE and LPIPS loss~\cite{lpips} at the input view and the additional view.  
Note that LRM is similar to DiT in that it injects camera information by modulating the image latents with adaptive layer norm (AdaLN).
LRM enables direct single-view to 3D reconstruction. The input image $I$ is encoded by an image encoder, producing patch-wise feature tokens $F \in \mathbb{R}^{N \times d_e}$, where $N$ is the number of image feature patches and $d_e$ is the dimension of the feature token. Initial learnable positional embeddings are added to the feature tokens and are fed into a transformer encoder to generate the triplane feature map $T$.


\subsubsection{Differentiable Marching Cubes (DiffMC)}
DiffMC uses the standard marching cubes algorithm to extract meshes and calculate vertex gradients $\frac{\partial v}{\partial g}$ for the grid, enabling smooth integration with mesh optimization via the chain rule:
\[ 
\frac{\partial \mathcal{L}_{ren}}{\partial \theta} = \sum_{v \in V} \frac{\partial \mathcal{L}_{ren}}{\partial v} \frac{\partial v}{\partial g} \frac{\partial g}{\partial \theta}, 
\] 
where \(\mathcal{L}_{ren}\) is the rendering loss, \(\theta\) represents the parameters in the density network, \(V\) is the set of mesh vertices, and \(g\) is the grid.
In addition, DiffMC introduces deformable vectors into the mesh for optimization, allowing the shape of each cube to be fine-tuned.
The 2D image rendering is performed using nvdiffrast, and the geometry and appearance networks are updated using the rendering loss. 
The surface points of the mesh are generated by the appearance network to generate pixel colors.
Compared with independent volume or mesh rendering, the pre-trained network can provide a powerful initialization function for volume rendering and traveling cubes, which is better than independent volume or mesh rendering.

\subsubsection{Diffusion Model}
Diffusion models (Sohl-Dickstein et al., 2015; Ho et al., 2020) aim to learn the probability distribution \( p_\theta(\mathbf{x}_0) = \int p_\theta(\mathbf{x}_{0:T}) \, d\mathbf{x}_{1:T} \). The reverse process is formulated as follows:  

\[  
p_\theta(\mathbf{x}_{0:T}) = p(\mathbf{x}_T) \prod_{t=1}^T p_\theta(\mathbf{x}_{t-1}|\mathbf{x}_t),  
\]  
where \( p(\mathbf{x}_T) = \mathcal{N}(\mathbf{x}_T; \mathbf{0}, \mathbf{I}) \) and \( p_\theta(\mathbf{x}_{t-1}|\mathbf{x}_t) = \mathcal{N}(\mathbf{x}_{t-1}; \mu_\theta(\mathbf{x}_t, t), \sigma_t^2 \mathbf{I}) \) is a parameterized Gaussian with a fixed variance \(\sigma_t^2\).
To learn \(\mu_\theta\), we construct a forward process defined by:  
\[  
q(\mathbf{x}_{1:T}|\mathbf{x}_0) = \prod_{t=1}^T q(\mathbf{x}_t|\mathbf{x}_{t-1}),  
\]
where \( q(\mathbf{x}_t|\mathbf{x}_{t-1}) = \mathcal{N}(\mathbf{x}_t; \sqrt{1-\beta_t}\mathbf{x}_{t-1}, \beta_t \mathbf{I}) \) and \(\beta_t\) are constants. 
The objective is to learn the denoising function \(\epsilon_\theta\), where the mean \(\mu_\theta\) is expressed as:  

\[  
\mu_\theta(\mathbf{x}_t, t) = \frac{1}{\sqrt{\alpha_t}} \left( \mathbf{x}_t - \frac{\beta_t}{\sqrt{1-\bar{\alpha}_t}} \epsilon_\theta(\mathbf{x}_t, t) \right).  
\]  

To learn \(\epsilon_\theta\), we optimize:  

\[  
\mathcal{L} = \mathbb{E}_{\mathbf{x}_0, \epsilon} \left[ \left\| \epsilon - \epsilon_\theta(\sqrt{\bar{\alpha}_t}\mathbf{x}_0 + \sqrt{1-\bar{\alpha}_t}\epsilon, t) \right\|^2 \right],  
\]  


\begin{figure*}[t!]
    \centering
    \includegraphics[width=1.0\textwidth]{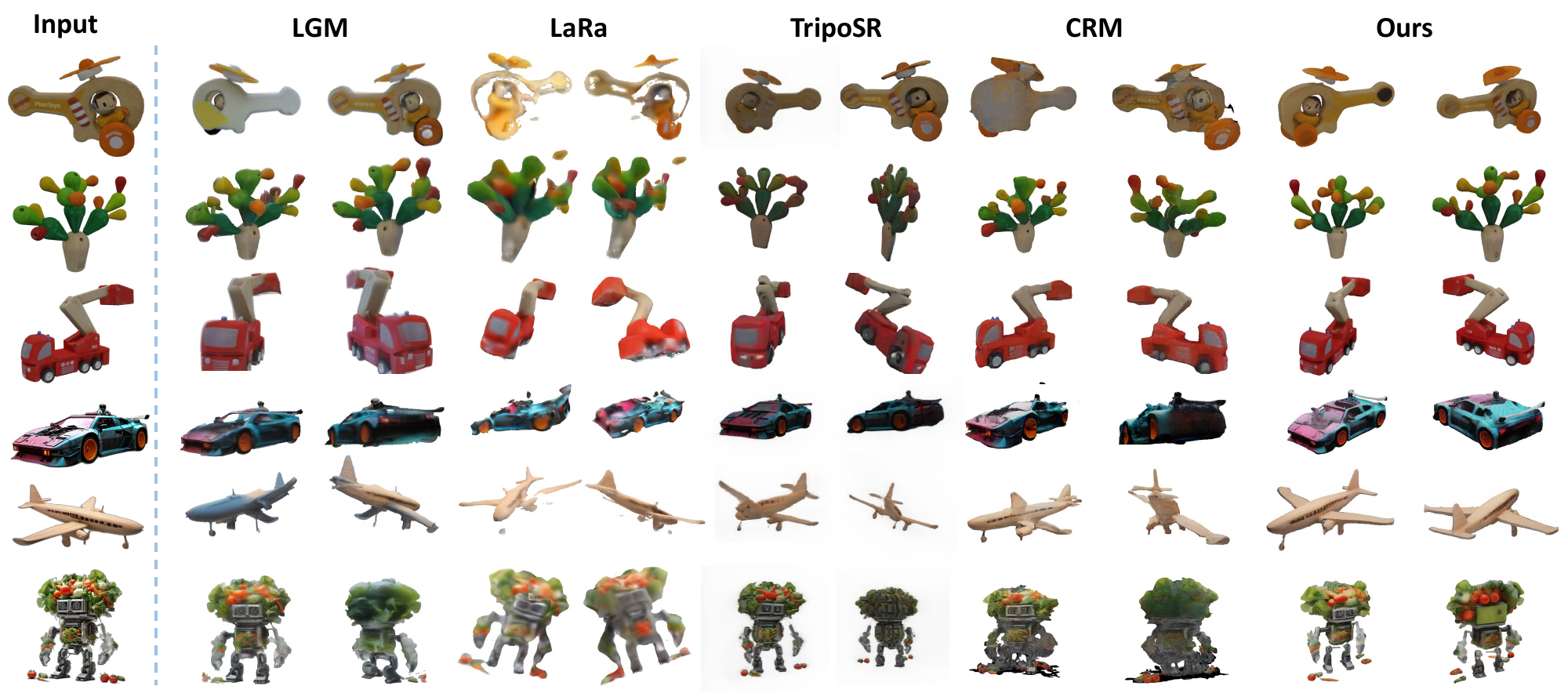}
    \vspace{-2mm}
    \caption{Qualitative comparison of single view 3D reconstruction results on the GSO and in-the-wild datasets. Though our method is not specifically designed for single-view input, it can achieve results comparable to these benchmark methods and generate more reasonable textures in the invisible area.}
    \label{fig:singleview}
    \vspace{-2mm}
\end{figure*}


\begin{figure}[t!]
    \centering
    \includegraphics[width=0.5\textwidth]{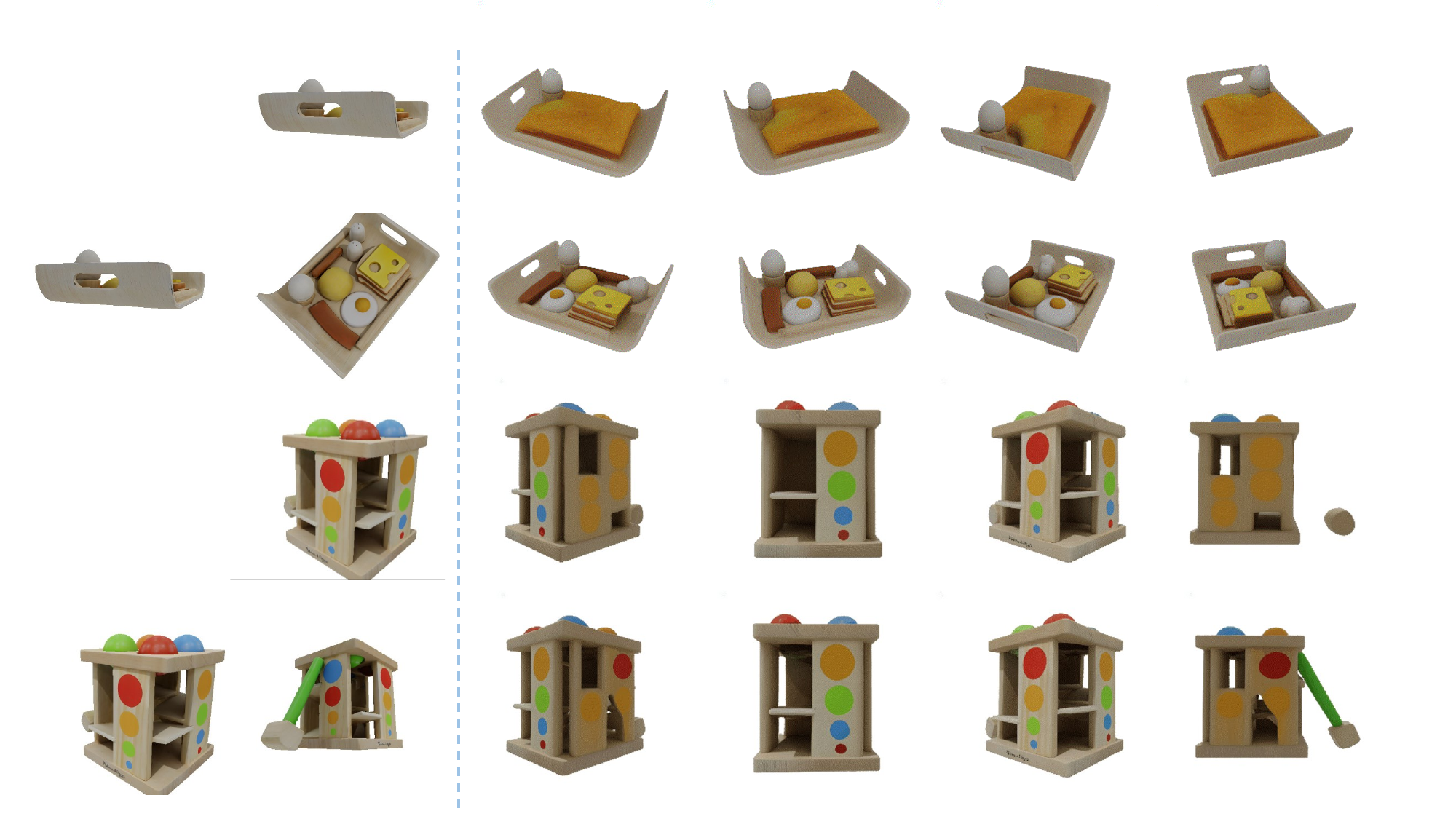}
    \vspace{-2mm}
    \caption{\textbf{Ablation Study on Generative Priors.} As shown above, incorporating generative priors enables the synthesis of plausible textures in unseen regions of objects.}
    \label{fig:onevsmulti}
    \vspace{-2mm}
\end{figure}
\begin{figure*}[t!]
    \centering
    \includegraphics[width=0.95\textwidth]{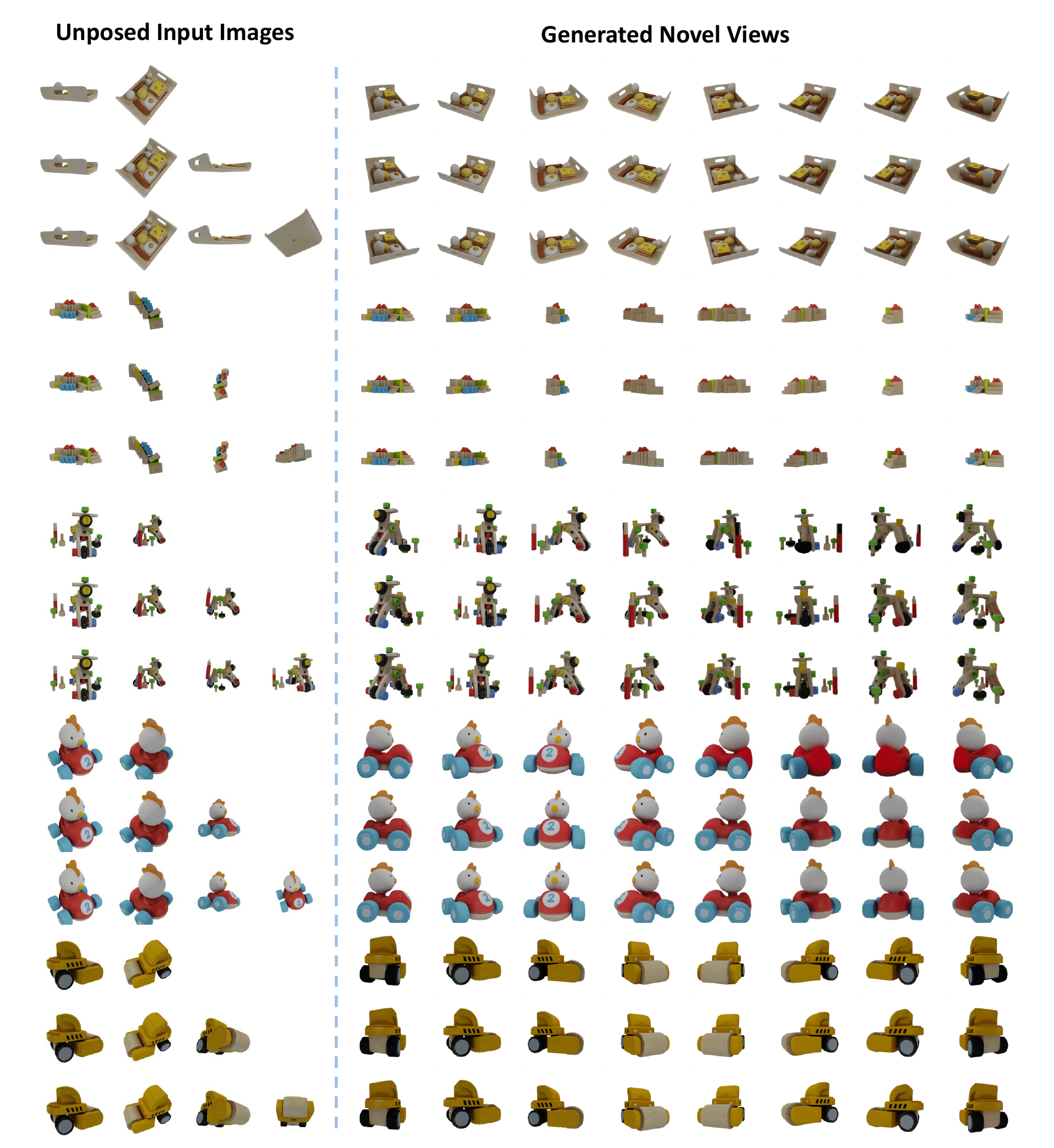}
    \vspace{-2mm}
    \caption{The visualization results of the novel view synthesis. The results show that our method can generate reasonable novel views from any number of input views, and the quality of the generated views improves with the number of input views.}
    \label{fig:multiview_gen}
    \vspace{-2mm}
\end{figure*}

\subsubsection{Multiview Diffusion Model}
Since the previous method separately applies the DDPM model to generate new view images, it is difficult to maintain consistency between different views. To solve this problem, Syncdreamer~\cite{liu2023syncdreamer} proposes to define the generation process as a multi-view diffusion model, which is associated with the generation of each view.
Let $\{\boldsymbol{x}_0^{i}\}_{i=1}^N$ denote the N images we aim to generate, where 0 indicates time step 0. 
The goal of the multiview diffusion model is to learn the joint distribution of all views.
The forward process of the multiview diffusion model is an extension of the origin DDPM:
\[  
q(\boldsymbol{x}_{1:T}^{(1:N)}|\boldsymbol{x}_0^{(1:N)})  = \prod_{t=1}^T \prod_{i=1}^N q(\boldsymbol{x}_t^{(i)}|\boldsymbol{x}_{t-1}^{(i)}),
\]  
Similarly, the reverse process is formulated as follows:
\[  
p_\theta(\boldsymbol{x}_{0:T}^{(1:N)}) = p(\boldsymbol{x}_T^{(1:N)}) \prod_{t=1}^T \prod_{i=1}^N p_\theta(\boldsymbol{x}_{t-1}^{(i)}|\boldsymbol{x}_t^{(i)})  
\]  
where $p_\theta^{(n)}(\boldsymbol{x}_{t-1}^{1:N}|\boldsymbol{x}_t) = N(\boldsymbol{x}_{t-1}^{1:N}; \mu_\theta^{(n)}(\boldsymbol{x}_t^{1:N}, t), \sigma_t I).$
Similar to DDPM, we define the mean and the loss by
\[  
\mu_\theta^{(n)}(\boldsymbol{x}_t^{1:N}, t) = \frac{1}{\sqrt{\alpha_t}} \left( \boldsymbol{x}_t^{(n)} - \frac{\beta_t}{\sqrt{1-\bar{\alpha}_t}} \epsilon_\theta^{(n)}(\boldsymbol{x}_t^{1:N}, t) \right)  
\]  
\[  
\mathcal{L} = \mathbb{E}_{t, \boldsymbol{x}_t^{1:N}, \epsilon} \left[ \| \epsilon^{(n)} - \epsilon_\theta^{(n)}(\boldsymbol{x}_t^{1:N}, t) \|_2^2 \right]  
\]  
where $\epsilon^{(1:N)}$  is the standard Gaussian noise, $\epsilon^{(n)}$ is the noise added to the n-th view, and $\epsilon_\theta^{(n)}$ is the noise predictor on the n-th view.

\subsection{A.2 Implementation details.}
Our model is optimized using the AdamW~\cite{adamw} with a weight decay of 0.01.
The multiview generator is trained with an initial learning rate of $3e-4$, which decreases to $1e-5$ over 30,000 training steps using a cosine learning rate decay schedule.
Gradient clipping is applied with a maximum norm of 1.0.
Similarly, the feed-forward reconstruction model employs a learning rate of $1e{-5}$, featuring cosine decay and a 1,000-step linear warm-up. Fine-tuning is conducted over 10,000 steps.  
The DiffMC~\cite{diffmc} operates at a resolution of 256 within a $[-1, 1]^3$ bounding box.
We trained on 16 A800 GPUs with a batch size of 64, using FP16 mixed precision and gradient storage, and using ZeRO-2 optimization to improve training efficiency and save GPU memory.



\subsection{A.3 More Results and Ablations}
\noindent{\textbf{Multiview Conditional Diffusion Model.}}
As shown in Figure~\ref{fig:multiview_gen}, our method can generate reasonable novel perspectives from any number of unposed input perspectives, and the quality of the generated perspective will improve as the number of input perspectives increases. This achieves a transformation from a random coordinate system to a canonical object coordinate system, thereby enabling a more accurate reconstruction of the given object. 

\noindent{\textbf{Single-View v.s. Multi-View.}}
As shown in Figure~\ref{fig:onevsmulti}, although the novel view generation based on a single view can generate reasonable results, there is a lot of stochasticity in the unseen areas, which may not be consistent with the texture of the real object.
Therefore, by introducing more control information through input multiple views, a high-fidelity reconstruction of the object can be achieved.

\subsection{A.4 Future Work}
Our method is different from recent multi-view diffusion models, which require a set of rendered images from a fixed viewpoint in a 3D dataset for training. 
Our method only requires a sparse set of unposed images for training. This makes it possible to train our model using large-scale video data, thereby further improving the object reconstruction ability and even enabling sparse view scene reconstruction. 
Therefore, extending our method to adapt to multiple data sources is an important direction for future research.

\end{document}